\documentclass{article}
\usepackage[bookmarks=false]{hyperref}

\usepackage{arxiv}

\usepackage[utf8]{inputenc} %
\usepackage[T1]{fontenc}    %
\usepackage{url}            %
\usepackage{booktabs}       %
\usepackage{amsfonts}       %
\usepackage{nicefrac}       %
\usepackage{microtype}      %
\usepackage[numbers]{natbib}
\usepackage{subcaption}
\usepackage[pdftex]{graphicx}
\usepackage{amsmath}
\usepackage{fancyvrb}
\usepackage{todonotes}
\usepackage{makecell}
\usepackage{cleveref}
\usepackage{authblk}

\graphicspath{ {./Figures/} }

\usepackage{siunitx}

\begin{document}
\bibliographystyle{unsrt}

\makeatletter
\renewcommand\AB@affilsepx{, \protect\Affilfont}
\makeatother

\renewcommand*{\Authfont}{\bfseries}

\title{Spectral Analysis of Latent Representations}
\author[1,2]{Justin Shenk}
\author[2,3]{Mats L. Richter}
\author[1]{Anders Arpteg}
\author[1]{Mikael Huss}
\affil[1]{Peltarion}
\affil[2]{University of Osnabrück}
\affil[3]{AIM Agile IT Managment GmbH}

\date{\vspace{-1.1cm}
\texttt{\{justin,anders,mikael\}@peltarion.com, matrichter@uos.de}
}

\maketitle

\begin{abstract}
We propose a metric, Layer Saturation, defined as the proportion of the number of eigenvalues needed to explain 99\% of the variance of the latent representations, for analyzing the learned representations of neural network layers.
Saturation is based on spectral analysis and can be computed efficiently, making live analysis of the representations practical during training.
We provide an outlook for future applications of this metric by outlining the behavior of layer saturation in different neural architectures and problems.
We further show that saturation is related to the generalization and predictive performance of neural networks.

\begin{figure}[h!]
\centering
    \begin{subfigure}{.3\linewidth}
        \captionsetup{width=\columnwidth}
          \centering
          \includegraphics[width=\columnwidth]{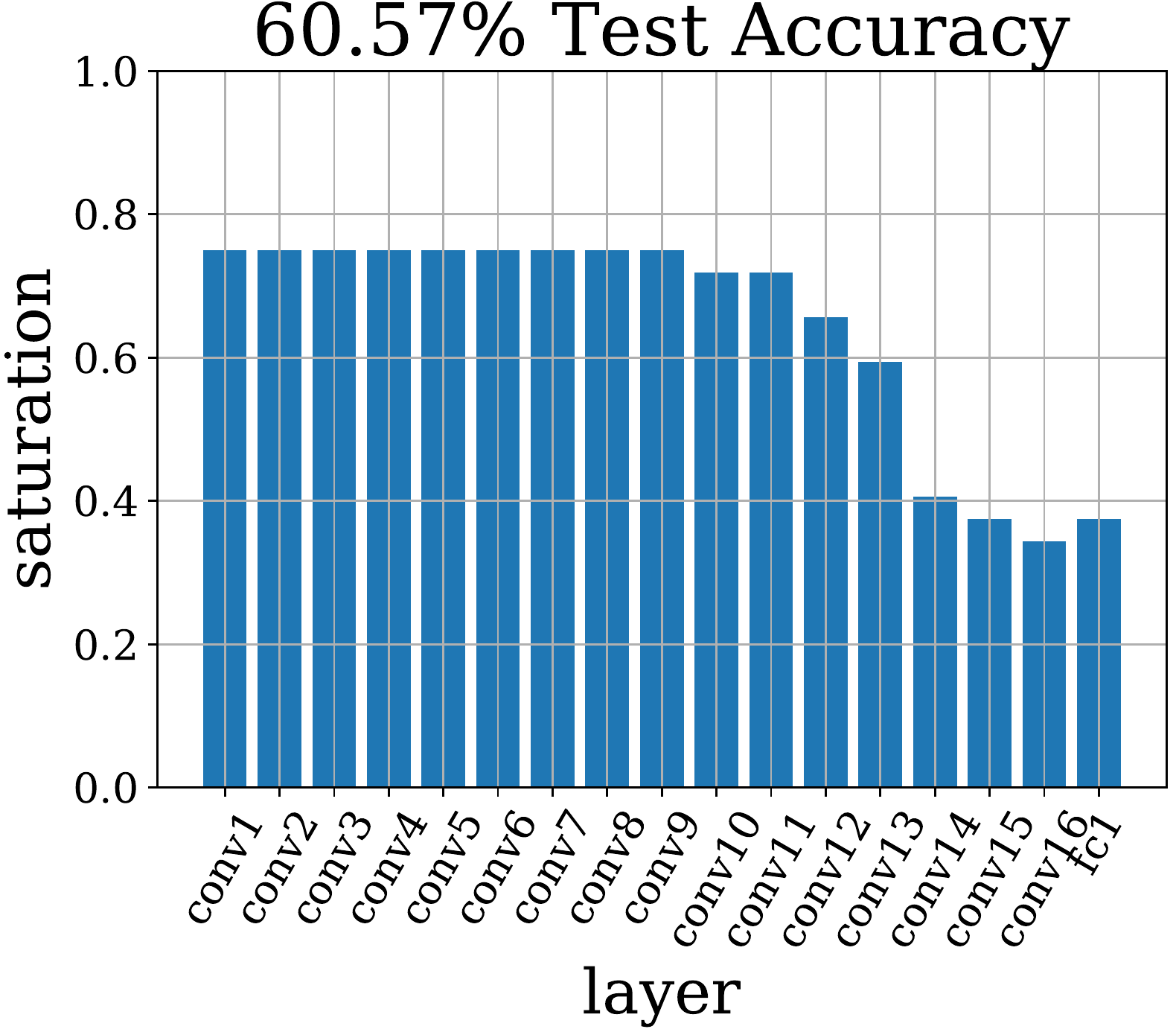}
         \caption[font=small]{not enough filters}
          \label{fig:to_little}
    \end{subfigure}
    \begin{subfigure}{.3\linewidth}
        \captionsetup{width=\columnwidth}
          \centering
          \includegraphics[width=\columnwidth]{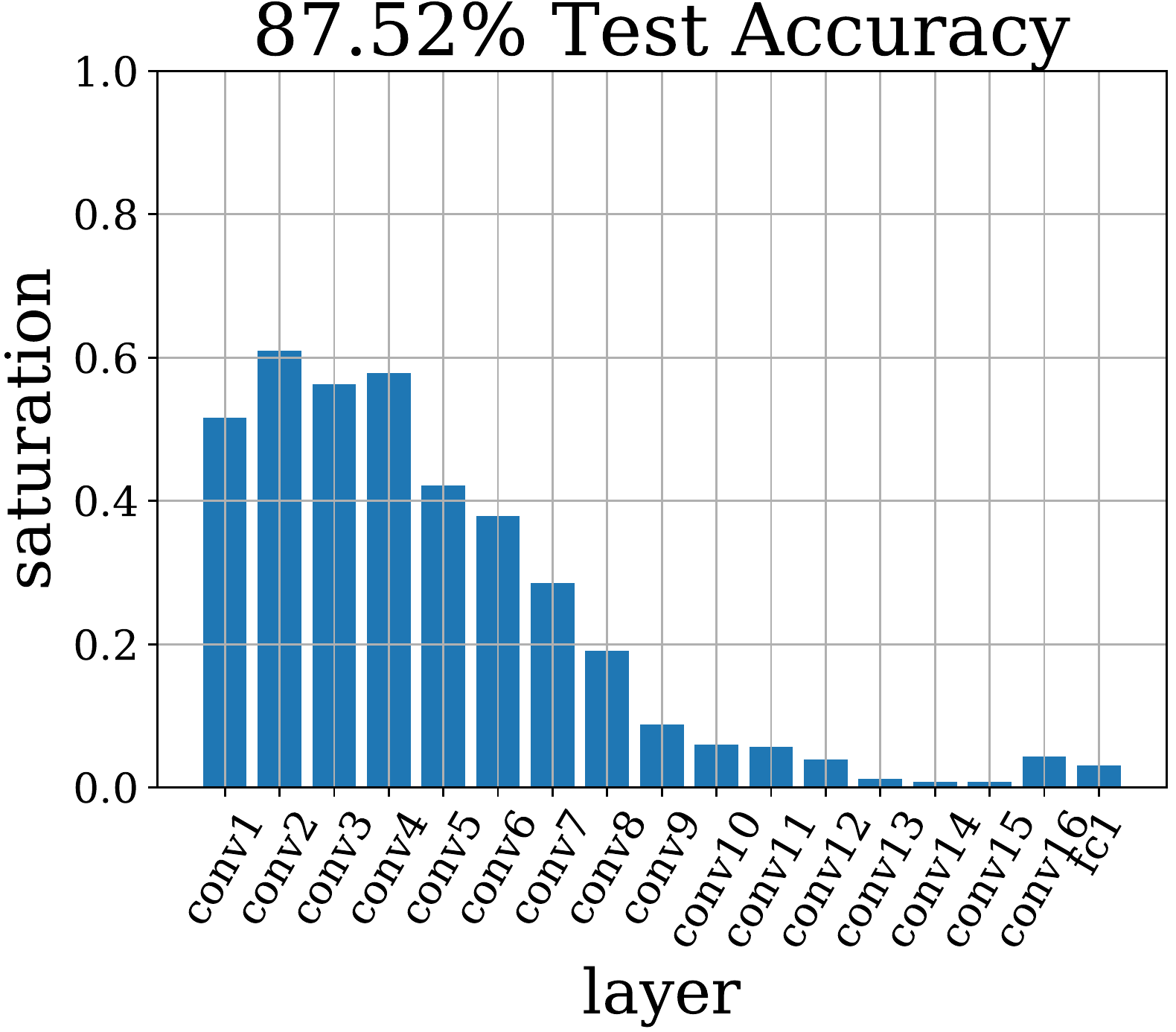}
          \caption[font=small]{too deep}
          \label{fig:just_right}
    \end{subfigure}
    \begin{subfigure}{.3\linewidth}
        \captionsetup{width=\columnwidth}
          \centering
          \includegraphics[width=\columnwidth]{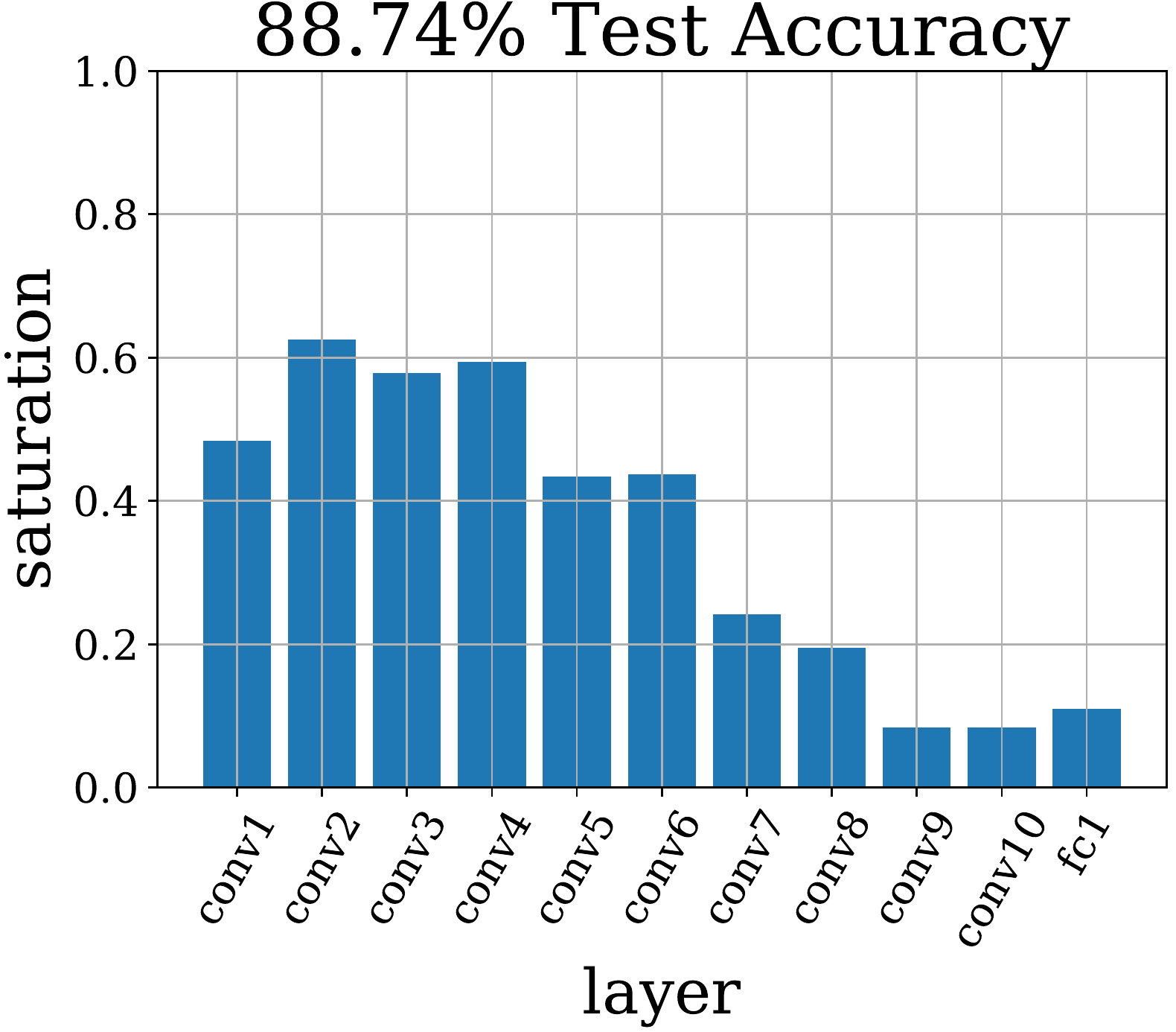}
          \caption[font=small]{just right}
          \label{fig:overparam}
    \end{subfigure}

    \label{fig:patterns}
    \caption{Layer-wise Saturation on three different VGG-style networks trained on CIFAR10.}
\end{figure}

\end{abstract}

\section{Introduction} %
\label{Introduction} %

Deep neural networks are very powerful but also opaque function approximators.
Opaque in the sense, that the transformation from high dimensional input data into an output over a sequence of latent representations is neither self-evident nor easily human readable \citep{alexnet, zfnet}.
Understanding the way information is processed through the network and how those representations change during training is important for developing a purpose-build neural network design process.

Multiple tools were developed over the last years in order to analyze neural network properties and behavior.
For instance \citep{keskar, sensitivitygoogle} proposed metrics for analyzing the sharpness of local optima and the stability of neural network predictions. Other publications like \citep{errorsurface} aim also at understanding the learning dynamics at training time as well as visualizing local parts of the loss landscape.
A method for analyzing and comparing learned representations of models was proposed by \citep{svcca}.
Even though these tools delivered new and interesting insights in deep neural networks, they are generally demanding computational tasks on their own.
For this reason it is impractical to integrate them into the training process of a model or to use them on-the-fly for additional insight.
We propose a simpler, fast to compute and easy to interpret metric, which enables the live insight into the evolving latent representation during training.
Like SVCCA, our method estimates the intrinsic dimensionality of layers via SVD/PCA, but in contrast to SVCCA, it can be calculated during training with just a few forward passes. Our approach is different from the aforementioned techniques because it can be calculated with only a few forward-passes at any point of training.

Saturation provides insight into the information processing on a layer-wise level during training.
This will provide the deep learning researcher with a fast to compute source of additional insight for understanding the behavior of the model.
We will demonstrate by experiments that saturation is showing interaction between generalization performance, model architecture and problem complexity.
We will further demonstrate that the metric may also be used to detect underparameterization in architectures and overfitting during training.

\section{Metric for understanding neural information processing}
The goals of the metric described in this paper are as follows:

\begin{itemize}
    \item Deliver insight into the flow of information in the neural network during training
    \item Make this information easy to compute and easy to interpret
    \item Guide model architecture selection
\end{itemize}

In order to achieve these goals, we analyse the learned representations of all layers inside the network.
The \textit{representation} of a neuron refers to its value before activation over a dataset.
By comparing the intrinsic dimensionality $i$ to the dimensionality of the feature space we can estimate how much the feature space is saturated by the latent representation of the input data.
In order to approximate the intrinsic dimensionality Principal Component Analysis (PCA) is performed on the latent representation matrix of fully connected and convolutional layers of a network.
From the pre-activation states $z_l$, where $l$ is a convolutional or fully connected layer, an autocovariance matrix $Cov_n(z_l, z_l)$ is computed on-line on every step $n$ during training using the following formula:

$$\frac{Cov_{n-1} \cdot (n - 1) + [ \frac{n-1}{n} ] (z_{l,n} - \bar{z}_{l,n-1})(z_{l, n} - \bar{z}_{l, n-1})}{n}$$

where $Cov_{n-1}$ is the covariance matrix of the previous timestep, $z_{l,n}$ is the latent representation of the layer $l$ at step $n$ and $\bar{z}_{l, n}$ the running mean of the latent representation of layer $l$ at step $n$. This iterative computation enables us to approximate the covariance matrices efficiently without keeping track of the entire history of latent representations and also spreading out the computation onto each training step.

We then compute the eigenvalues $\lambda$ of those covariance matrices.
The eigendirections  responsible for 99\% of the variation can be obtained by summing up the eigenvalues $\{\lambda_1,...,\lambda_{m1}\}$ in descending order until 99\% of the sum of all eigenvalues is reached.
The number of those eigendirections $m_{1}^{\prime}$ is our approximation of the intrinsic dimensionality $i$ of layer $l$. We extend the nomenclature introduced in \citep{svcca} by defining layer saturation as
\begin{equation}
s = \frac{m_{1}^{\prime}}{|l|} = \frac{\sum_{i=1}^{m_{1}^{\prime}} 1 (\geq 0.99 \sum_{i=1}^{m_1} |\lambda_i|)}{|l|}
\end{equation}

where $|l|$ is the size of the representation in each layer with $s \in {\mathbb{R}} : 0 \leq s \leq 1 $.

For example, in the case of a 2D convolutional layer containing 32 feature maps, the average value for each of the 32 feature maps is used to calculate the intrinsic dimensionality and saturation (more on that later).
In the case of a fully-connected layer of 64 neurons, the deep representations of all 64 neurons are accumulated and analyzed.
The space spanned by the corresponding eigenvectors can be interpreted as an approximation of the subspace that the input data is mapped on.
Thus the approximated intrinsic dimensionality is the number of eigenvalues $m_{1}^{\prime}$ and the saturation is $\frac{m_1^\prime}{|l|}$.

\subsection{Model Average Saturation}
Deriving information about the whole architecture is difficult using layer-wise saturation, due to the large number of saturation values.
In order to keep the interpretability easy for inferring properties of the entire architecture, a simplification is required.
The \textit{average saturation} $s_{\mu}$ for a model is a derived metric from layer saturation.
It is defined as follows:

\begin{equation}
    s_{\mu} = \frac{1}{|D|} \sum^{D}_{l} s_l
\end{equation}
with $D$ as the set of non-output fully connected and convolutional layers in the network and $s_l$ the saturation of layer $l$. We will show \textit{average saturation} is behaving similar to layer-wise saturation when making changes to the neural architecture.
We will further demonstrate that, similar to saturation on a single-layer, average saturation can be used to detect underparameterization.
We will further show that saturation is influenced by the depth as well as the general width of the network.

\subsection{Saturation for Convolutional Layers}

Since convolutional layers are agnostic towards the shape of their input- and output-tensors, computing saturation on the latent representation for a convolutional layer in any network is more challenging.
In order to achieve this, the latent representation needs to be projected into a space that has the same shape regardless of the height $H$ and width $W$ of the latent tensors $\mathbf{Z}^l$ of convolutional layer $l$.

Global pooling strategies are used as a common feature in neural architectures for a similar purpose \cite{nin, vgg, resnet}.
The most common global pooling strategy is global average pooling (GAP).
GAP computes the mean $\bar{\mathbf{Z}_{l,c} }$ of each filter $\mathbf{Z}_{l,c} $ of the latent representation $\mathbf{Z}_l$.
Global pooling effectively reduces a $H \cdot W \cdot C$-tensor with arbitrary positive integer values for $H$ and $W$ into a vector of length $C$ with $C$ being the number of filters.
\begin{equation}
    \mathbf{\bar{Z}}_{l,c}=\tfrac{1}{H_{\mathbf{Z}_l}\cdot W_{\mathbf{Z}_l}}\sum_h\sum_w \mathbf{Z}_{l,h,w,c}
\end{equation}
Since $C$ is a fixed value for any convolutional layer, the saturation of the pooled latent tensor $\bar{\mathbf{Z}_l}$ can be computed agnostic of $H_{\mathbf{Z}_{l,c} }$ and $W_{\mathbf{Z}^{l,c} }$.
The assumption made by using global average pooling is that each filter in the latent representation represents an activation map of a specific feature and all features are evenly activated over the image.
These assumptions are not necessarily correct.
However, the successful application of GAP as a simple untrained architectural feature in publications like \citep{vgg, resnet} shows, that critical information is maintained to a sufficient degree, despite the drastic reduction in dimensionality

\section{Experiments} %
Multiple experiments were conducted on CIFAR10 in order to understand the behavior of layer saturation under different circumstances.
The first experiments focus on per-layer saturation.
The primary purpose of those experiments is to understand the behavior  of the metric on a per-layer basis, e.g. how it is affected by changes of the layer and overfitting.
The final set of experiments will focus on saturation in large architectures.
We will first look into the properties of average saturation in large architectures and then analyze some examples in greater detail.

\subsection{Details about the experiments}
All experiments were performed using the ADAM optimizer and a batch size of 128.
For data preprocessing channel wise normalization based on image-net values was used.
Data augmentation during training consisted of random cropping with 4-pixel zero padding and horizontal flipping.
The first set of experiments, conducted on an convolutional autoencoder based on InceptionV3, was trained for 60.000 steps.
Training on convolutional and fully connected architectures in later experiments was performed for 20 epochs regardless of dataset, network and architecture.
All experiments were repeated multiple times in order to rule out the influence of factors like weight initialization or batch sampling. The results displayed in the scatterplots \Cref{fig:loss_sat_comp,fig:complexity_cifar,fig:test_acc_scatter,fig:scatter_conv_test_acc,fig:cnn_losses} are aggregated over all experiments and repetitions.

The CIFAR10 dataset \cite{CIFAR-10} consists of 60.000 train and 6.000 test images belonging to 10 classes.
The images are 32x32 pixel RGB-images.
For the experiments we wanted an an additional, simpler problem which is also related to CIFAR10, so we could draw conclusions about how problem difficulty affects the saturation of the network.
However, the problem must not be trivially solvable in order to still yield to useful experimental results.
The \textit{cat-vs-dog} dataset is a subproblem of CIFAR10 and was used for this purpose.
It only contains the cat and dog classes, thus having only 12,000 training and 1,200 test images.
The classes \textit{cat} and \textit{dog} were chosen since they are from our observation the most often confused pair of classes, when looking at the results of most classifiers we trained on CIFAR10.

In order to validate our findings in the final set of experiments with the VGG-architectures we also used another CIFAR10-derived dataset, we call \textit{unbalanced-CIFAR10}.
This dataset features a random imbalance between the 10 classes.
The least represented class makes up only 5\% of the sampled data, while the most sampled class is featured 15\% of time.
The remaining classes are interpolated in their sampling rate between those two.
The imbalance is randomly assigned to the classes for each experiment.

The convolutional autoencoder utilizes InceptionV3-architecture \cite{inceptionv3} as an encoder.
The data is up-sampled to $299\times 299\times 3$ in order to fit the data in the model.

The fully connected neural networks consist of three layers with an 128 unit input layer, an hidden layer of 8, 16, 32, 64 or 128 units and a softmax layer with 10 or 2 units. The first two layers use ReLU activation functions.

The VGG-style architectures contain the convolutional part of VGG11, 13, 16 and 19, followed by Global Average Pooling and a single fully connected layer with a number of units equal to the filter size of the last convolutional layer and a softmax.
Besides the alteration in depth, five variations in \textit{width} were used for each depth level.
These alterations have the full, half, quarter, eighth or sixteenth of the original number of filters in each of their layers.
This sums up to 20 convolutional architectures for this specific set of experiments.
We did this, since convolutional neural networks effectively have two commonly used degrees of freedom when it comes to altering the network complexity and we wanted to explore both of them.

\subsection{Convolutional Autoencoder}

\begin{figure*}[h!]
\centering
    \begin{subfigure}[t]{.5\textwidth}
          \centering
        \includegraphics[height=1.6in]{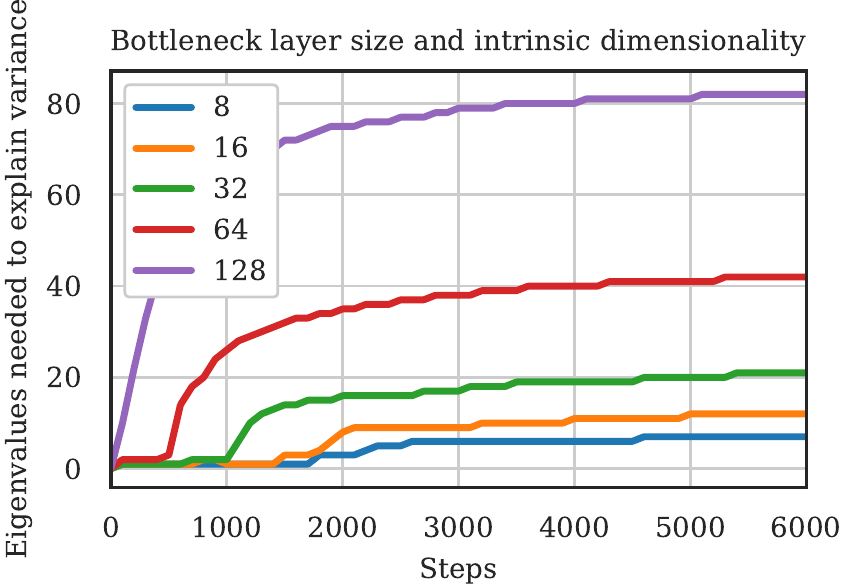}
          \caption[size_vs_intrinsic]{Intrinsic dimensionality increases with the number of units in the bottleneck layer.}
          \label{fig:size_vs_intrinsic}
    \end{subfigure}%
    ~
    \begin{subfigure}[t]{.5\textwidth}
          \centering
        \includegraphics[height=1.6in]{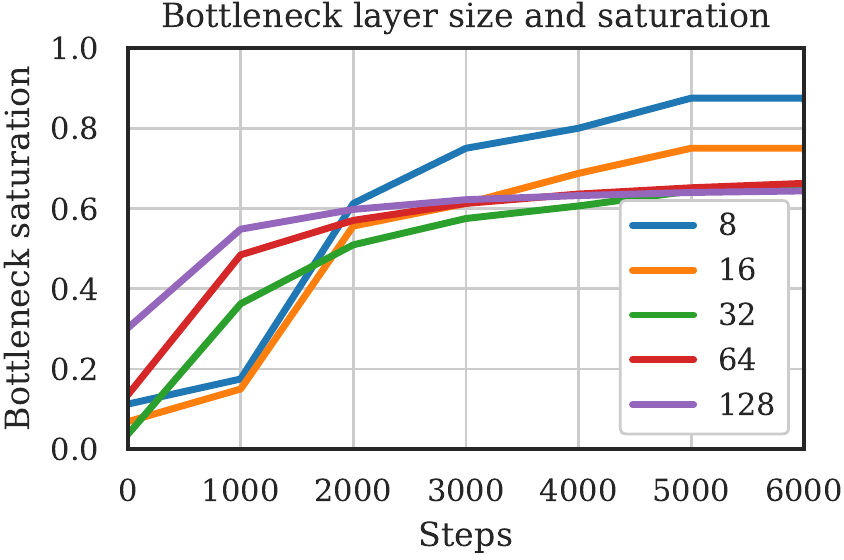}
          \caption[intrinsic_dims]{Saturation computed on regular steps during training. Note that wider bottleneck layers have higher capacity than narrow layers.}
          \label{fig:size_vs_sat}
    \end{subfigure}
    \caption{Wider layers yield higher intrinsic dimensionality but also lower saturation (CIFAR-10).}
    \label{fig:ae_sat_intrinsic}
\end{figure*}

The first experiments focus on the saturation of the bottleneck layer of a convolutional autoencoder, which is a good target for evaluating the properties of intrinsic dimensionality.
Autoencoder architecture was selected because the bottleneck layer is frequently used in practice for compressing input data. The relationship between the input data and the learned representations in the bottleneck are a good target for approaches to optimizing layer size.
As \Cref{fig:size_vs_intrinsic} shows, the bottleneck's size and the intrinsic dimensionality are related to each other.
Bottlenecks with higher neuron count have a higher intrinsic dimensionality. However, \Cref{fig:size_vs_sat} shows that this growth is not necessarily proportional to the increase in feature-space dimensionality induced by the higher bottleneck-layer neuron count, with wider bottlenecks having a lower saturation than narrower layers.

\begin{figure*}[t]
\centering
    \begin{subfigure}[t]{.5\textwidth}
          \centering
        \includegraphics[height=1.6in]{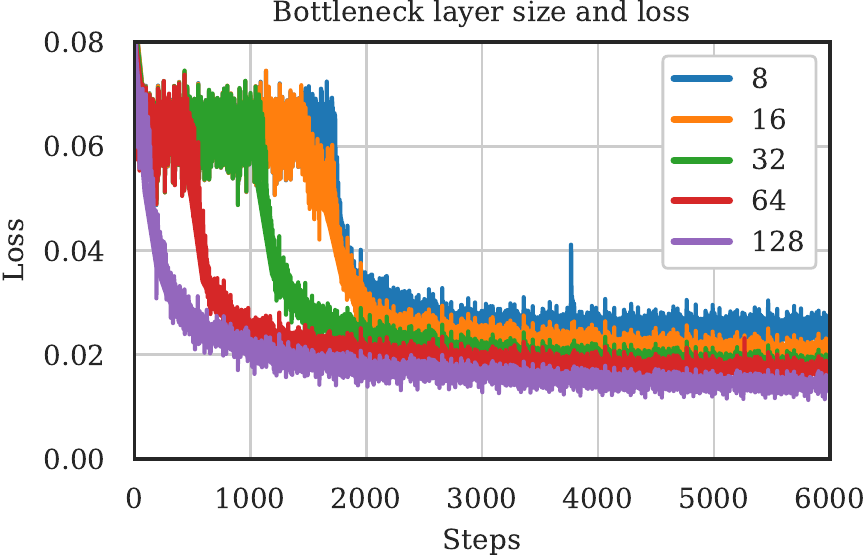}
          \caption[train_loss]{Train loss.}
          \label{fig:layersize_vs_train_loss}
    \end{subfigure}%
    ~
    \begin{subfigure}[t]{.5\textwidth}
          \centering
        \includegraphics[height=1.6in]{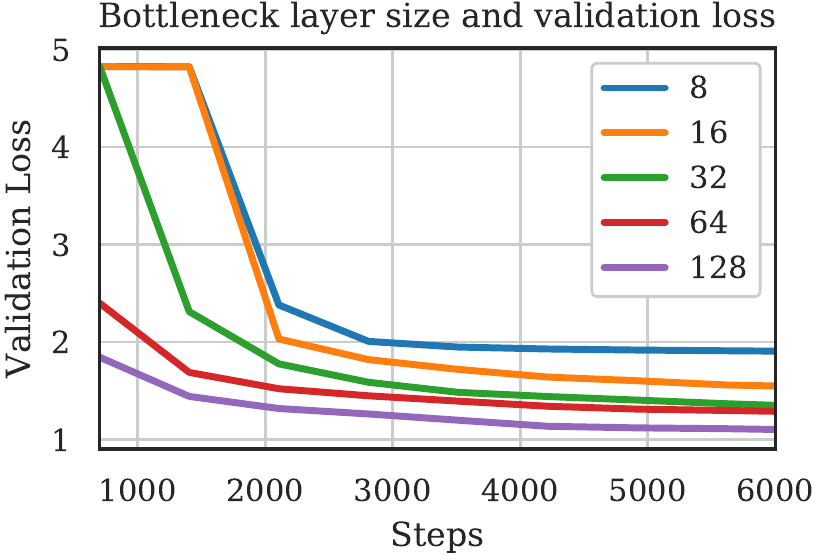}
          \caption[size_vs_vloss]{Validation loss during training.}
          \label{fig:layersize_vs_val_loss}
    \end{subfigure}
    \caption{Saturation and validation loss are inversely proportional during training (CIFAR-10).}
    \label{fig:loss_sat_comp}
\end{figure*}

Another interesting observation is, that especially the layer saturation is proportional to the validation loss, as can be seen in \Cref{fig:ae_sat_intrinsic,fig:loss_sat_comp}.
We interpret this as a symptom of the convergence of the model towards a local minimum.
High increase in intrinsic dimensionality suggest an ongoing learning process, since the manifold that the latent representations are living in is changing constantly.
In later, more converged parts of the training process the manifold in the feature space containing the latent representations of the data becomes increasingly stable and thus does no longer increase the saturation.

\subsection{Layer-wise saturation in classification problems and overfitting}
We conducted multiple experiments on dense neural networks, in order to reproduce and validate the findings of the autoencoder experiments as well as looking into the relationship between loss, accuracy and average saturation.
For this, we first ran a set of experiments on CIFAR10 and varied the number of neurons in the hidden layer of three-layer fully connected neural networks.

\begin{figure}[htb!]
\centering
    \begin{subfigure}{.5\textwidth}
          \centering
          \includegraphics[height=1.6in]{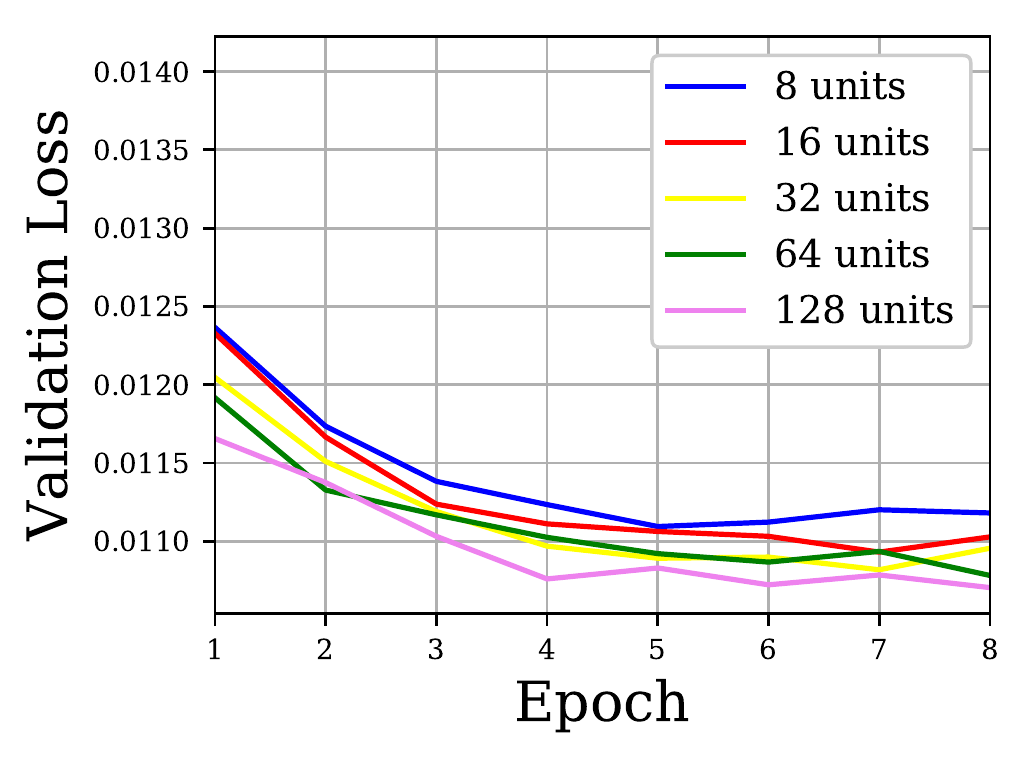}
          \caption[size_vs_vloss]{Validation loss during training.}
          \label{fig:loss_curve_dense}
    \end{subfigure}%
    ~
    \begin{subfigure}{.5\textwidth}
          \centering
          \includegraphics[height=1.6in]{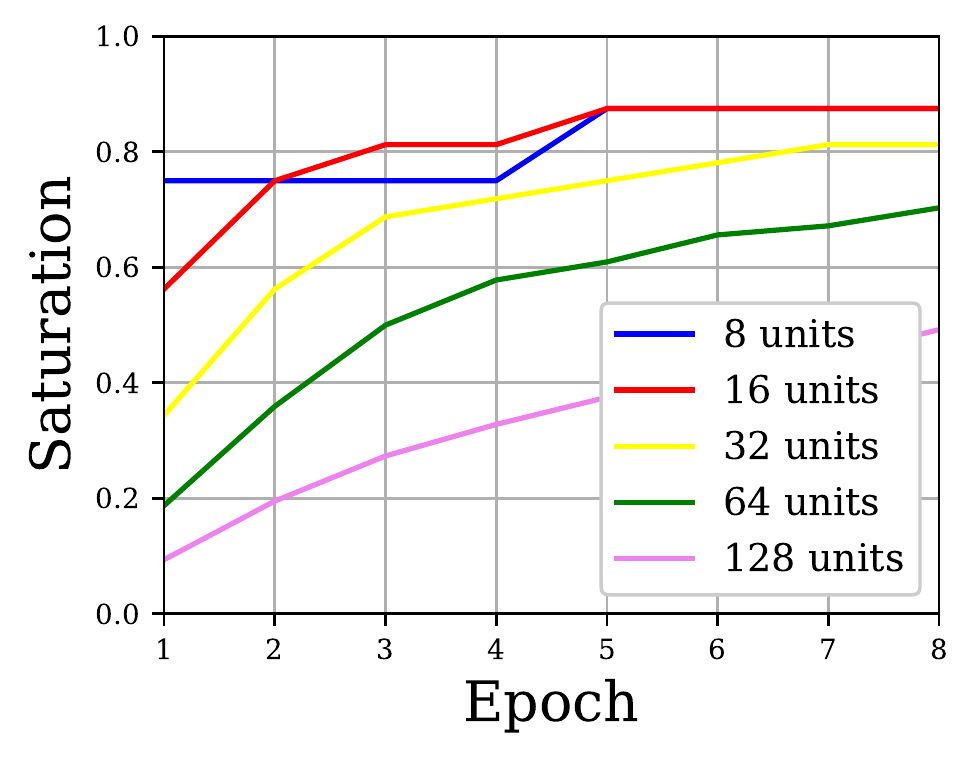}
          \caption[intrinsic_dims]{Saturation of layer 2 during training.}
          \label{fig:sat_curve_dense}
    \end{subfigure}
    \caption{Saturation and loss are behaving proportional during training, similar to the observed behavior for the bottleneck layer in \Cref{fig:loss_sat_comp}.}
    \label{fig:learning_curves_dense}
\end{figure}

As can be seen in \Cref{fig:learning_curves_dense}, the behavior of saturation during training shows similar behavior to the encoding layer of the auto encoder.

In order to look further into the relation of saturation and predictive performance, we repeated the experiment multiple times on CIFAR10 as well as on the simpler \textit{cat-vs-dog} binary-subproblem of CIFAR10.

\begin{figure}[htb!]
\centering
\captionsetup{width=\columnwidth}
\centering
\includegraphics[scale=0.8]{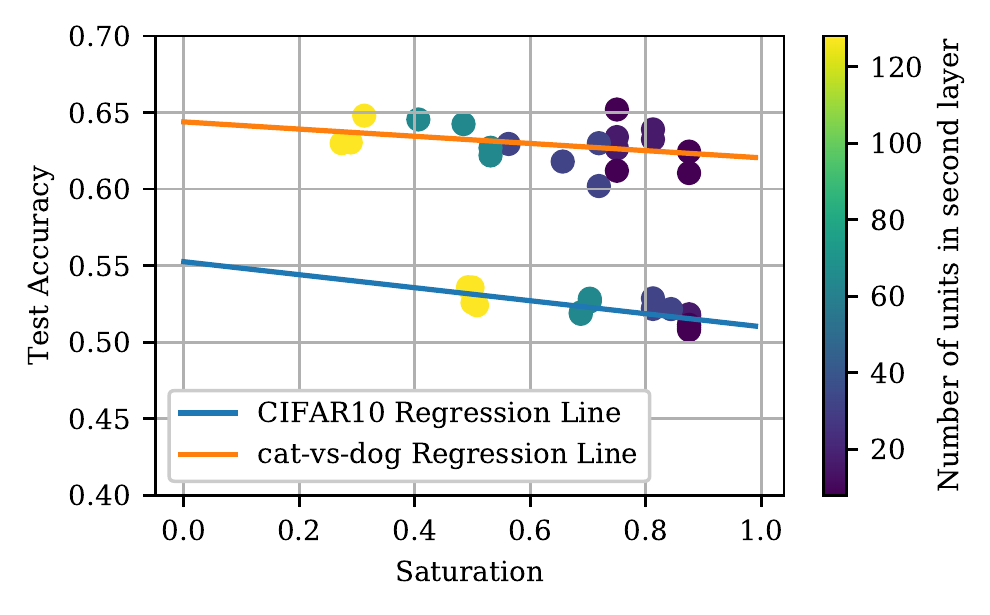}
\caption[size_vs_loss]{Saturation of the 2nd layer after training versus test accuracy.}
\label{fig:test_acc_scatter}
\end{figure}

We can see in \Cref{fig:test_acc_scatter} the higher parameterized networks generally achieve better performance and lower saturation levels, when compared to networks trained on the same problem but with fewer parameters.
When trained on problems of different complexity, the saturation is higher for the more complex problem given the same neural architecture and training conditions as more capacity is taxed by the more complex task.
Regardless of the problem however, the behavior of the saturation stays the same.
Performance of lower saturation models tends to be better than with higher saturation models, until overfitting occurs.
This is useful for model comparison purposes, since the saturation is bounded between 0 and 1, saturation values are not only indicative of model performance, but also whether the dimensionality of the layer and (as we will see later) the network is sufficient.

\begin{figure*}[htb!]
\centering
    \begin{subfigure}{0.5\textwidth}
          \centering
          \includegraphics[height=1.6in]{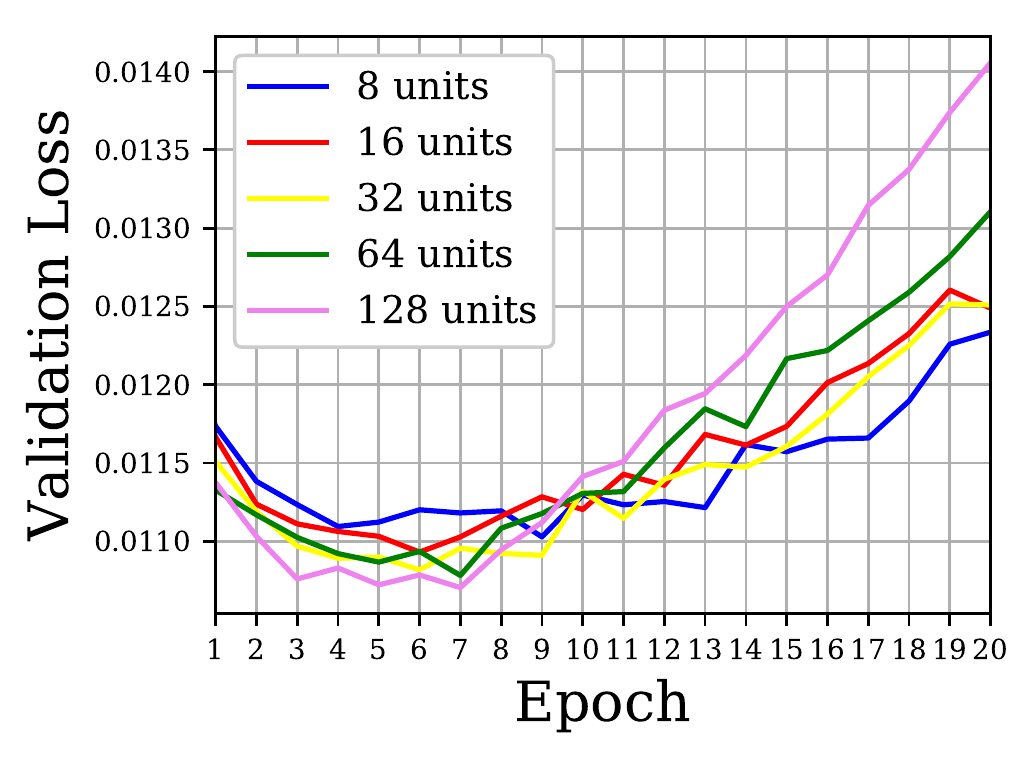}
          \label{fig:overfit_loss}
    \end{subfigure}%
    ~
        \begin{subfigure}{0.5\textwidth}
          \centering
          \includegraphics[height=1.6in]{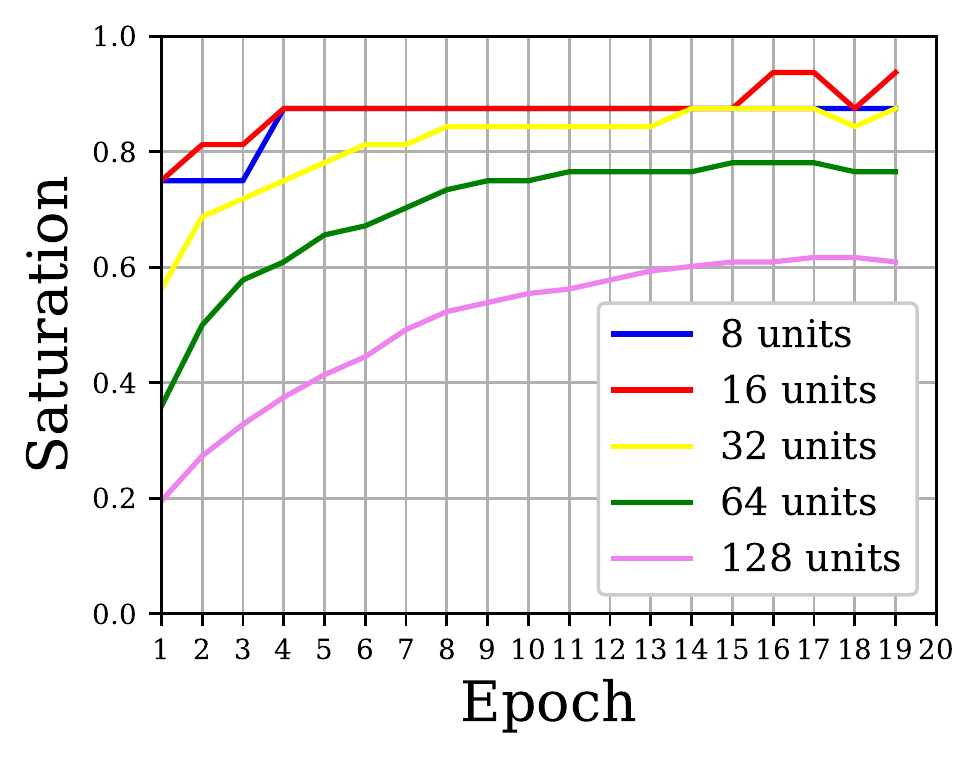}
          \label{fig:overfit_sat}
    \end{subfigure}
    \caption{Saturation of Layer 2 (left) and the validation loss (right) of overfitting fully connected networks trained on CIFAR 10. Saturation changes inversely with validation loss until the model starts overfitting.}
    \label{fig:overfit_loss}
\end{figure*}

When the network starts overfitting on a problem, the validation loss stop being proportional to the saturation as \Cref{fig:overfit_loss} shows.

\begin{figure}[t]
\centering
\captionsetup{width=\columnwidth}
\centering
\includegraphics[scale=0.6]{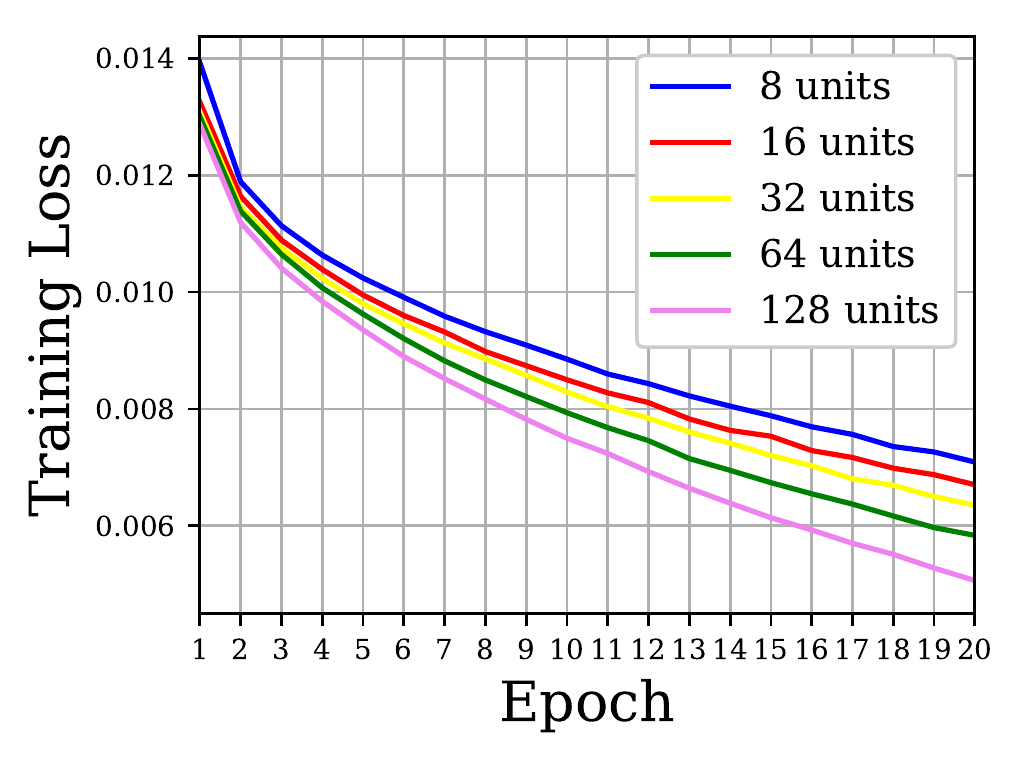}
\caption[sizevsloss]{Training loss during 20 epochs of training, note the trajectory seems to be in an earlier state of convergence than the curve of the saturation at the same epoch.}
\label{fig:overfit_train_hist}
\end{figure}

Interestingly, the training loss shown in \Cref{fig:overfit_train_hist} seems to be slower converging than the layer saturation, which indicates that the dimensionality of the latent representation becomes less volatile at first while the more detailed structure of the bounding manifold of the latent representation is still changing.
This property could be used in future work for developing a more sophisticated early stopping mechanism, that takes the model's state into account.

\begin{figure}[h]
\centering
\captionsetup{width=\columnwidth}
\centering
\includegraphics[scale=0.8]{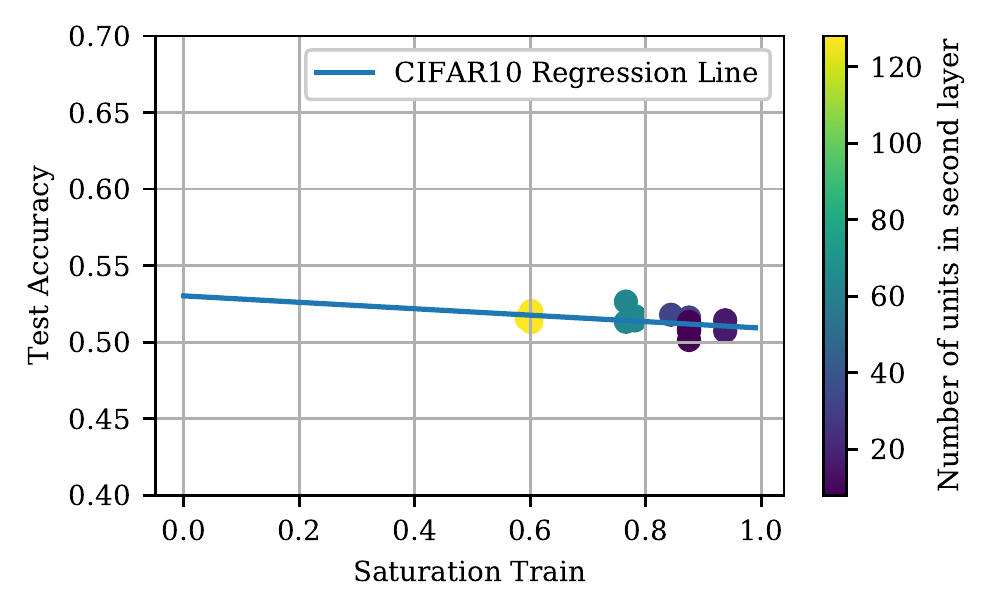}
\caption[sizevsloss]{Saturation of the 2nd layer versus test accuracy after 20 epochs of training.}
\label{fig:overfit_hist}
\end{figure}

However, the correlation between saturation and accuracy still remains as \Cref{fig:overfit_hist} shows, with higher neuron-count layers having lower saturation and a tendency to better performance.

This is interesting, since this means that overfitting is not related to the dimensionality of the bounding latent manifold but rather a structural difference inside the bounding latent manifold, which is not directly related to its dimensionality.

\subsection{Average Saturation of deep architectures}

\begin{figure}[t]
\centering
    \begin{subfigure}{0.5\textwidth}
          \centering
          \includegraphics[height=1.6in]{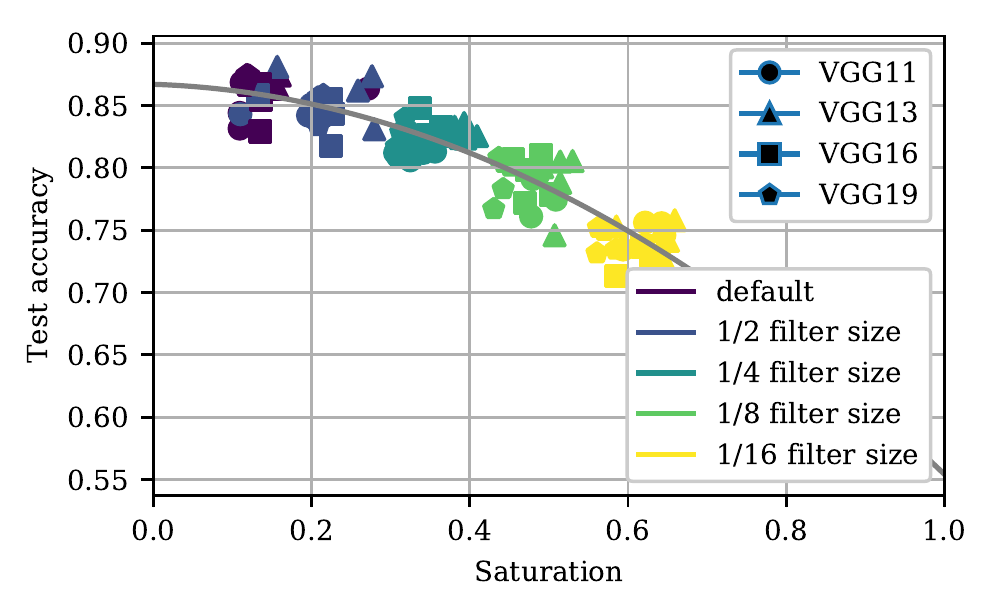}
          \caption[intrinsic_dims]{Cat vs Dog}
          \label{fig:complexity_catdog}
    \end{subfigure}%
    ~
        \begin{subfigure}{0.5\textwidth}
        \captionsetup{width=\columnwidth}
          \centering
          \includegraphics[height=1.6in]{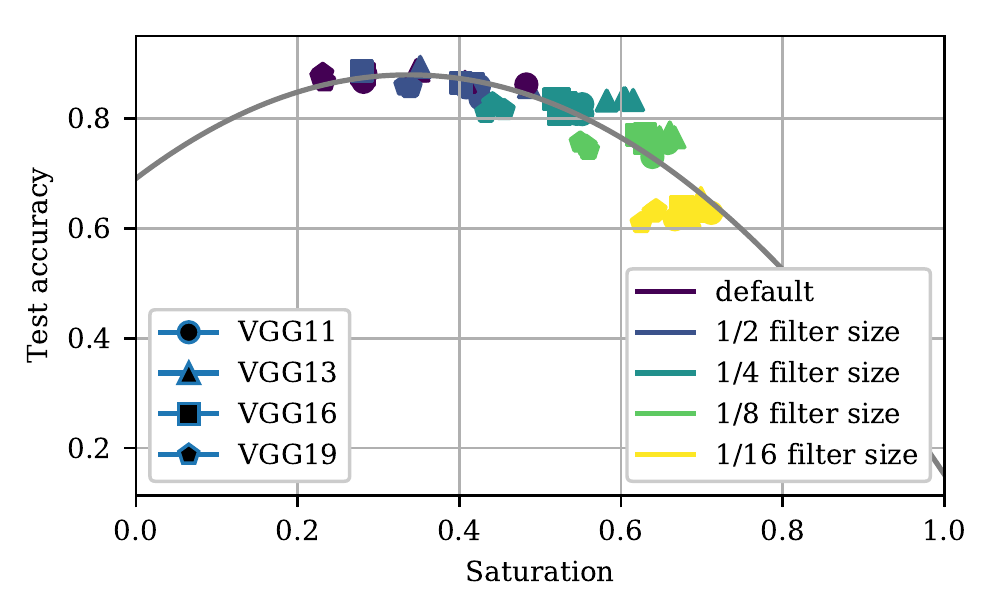}
          \caption[intrinsic_dims]{CIFAR10}
          \label{fig:overfit_sat}
    \end{subfigure}
    \caption{On both classification problems saturation is influenced by network depth and number of filters per layer. However, the number of filters generally has a more severe impact on saturation and performance.}
    \label{fig:complexity_cifar}
\end{figure}

For the last set of experiments we focused our analysis on the saturation of large and deep architectures.
The neural architectures were based on the VGG-family of neural networks with varying depth (11, 13, 16 and 19 layers) and varying filter sizes (from full original filter stack sizes to a sixteenth in exponential steps).

The general observation of lower saturation results in higher performance is also true for average saturation on large convolutional networks as can be seen in \Cref{fig:complexity_cifar}.
However, the pattern shown is quite interesting, as the performance decreases more rapidly for the more complex problem after exceeding roughly 45\% saturation.
This suggests, that average saturation can be viewed as an indicator whether the network used is complex enough to process the problem in a satisfying manner.
The saturation itself is influenced by the depth of the network as well as by the size of the filter-stacks of each layer, which can be seen in \Cref{fig:complexity_cifar}.

\subsection{Layer-wise saturation of deep architectures}

\begin{figure*}[h!]
\centering
    \begin{subfigure}{.4\linewidth}
        \captionsetup{width=\columnwidth}
          \centering
          \includegraphics[width=\columnwidth]{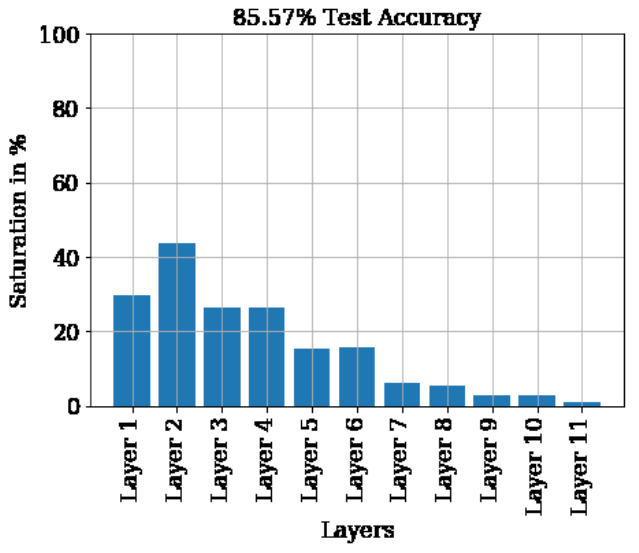}
          \caption[cvd]{Cat vs Dog}
          \label{fig:simple_problem}
    \end{subfigure}
    \begin{subfigure}{.4\linewidth}
        \captionsetup{width=\columnwidth}
          \centering
          \includegraphics[width=\columnwidth]{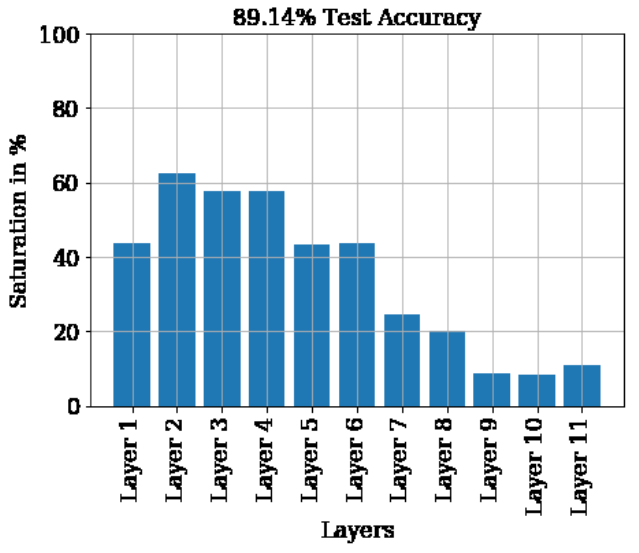}
          \caption[cif10]{CIFAR10}
          \label{fig:complex_problem}
    \end{subfigure}
    \caption{VGG13 trained on \textit{cat-vs-dog} is less saturated on a layer-wise level than on CIFAR10. This shows that saturation is not only a property of the architecture but also dependent on the complexity of the problem at hand.}
    \label{problem_complexity}
\end{figure*}

When analyzing the layer wise saturation, we can see patterns in the distribution of saturation.
The prototypical pattern of a well-fitted model can be seen in \Cref{fig:complex_problem}, where the saturation starts on high values in early layers.

\begin{figure}[htb!]
\centering
        \captionsetup{width=\textwidth}
          \centering
          \includegraphics[width=\textwidth]{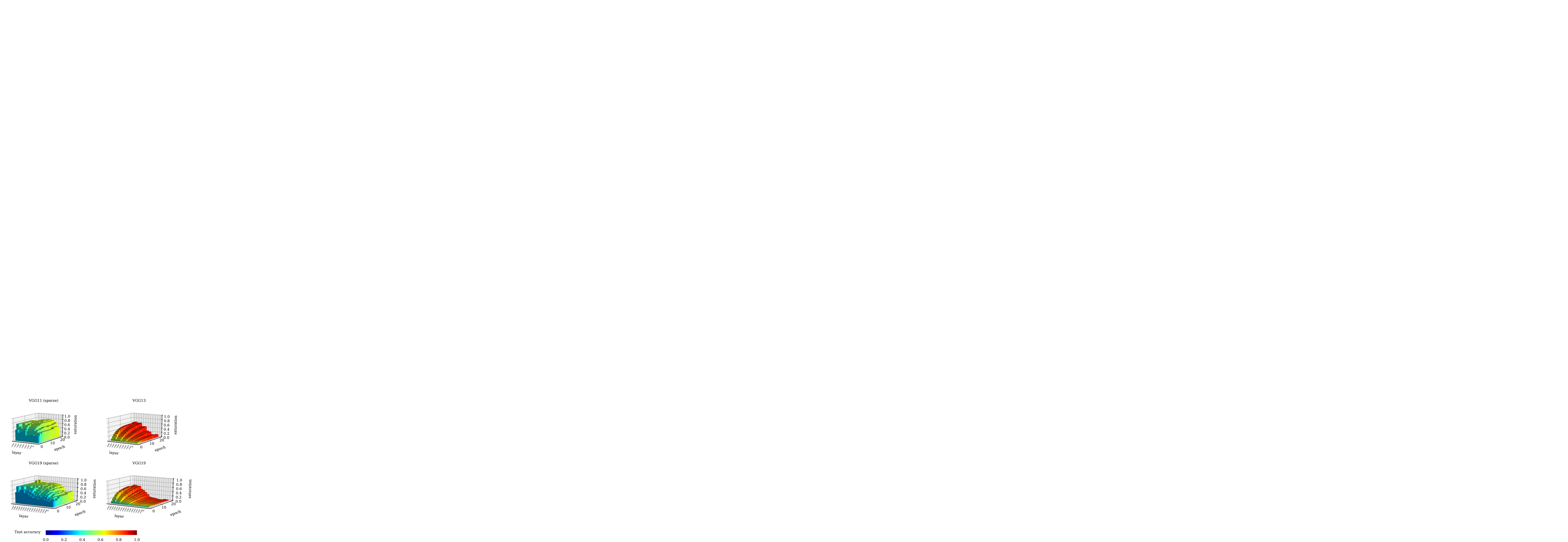}
          \caption[3d]{The evolution of saturation shows characteristic behaviors of over and underparameterized networks. Overparameterized networks (bottom right) evolve a longtail-like distribution of saturations. Underparameterized networks tend to have a large saturation from the very beginning (top left, bottom left).}
          \label{fig:sat_3d}
\end{figure}

The saturation rises shortly in the next couple of layers until it starts decreasing and eventually hits relatively stable level.
The length of this low-saturation long tail is dependent on the properties of the network, as can be seen in Figure 1.
The layer saturations of the network depicted in \Cref{fig:overparam} (VGG19) form a long tail due to overparameterization (too deep and too many filters).
The network depicted in \Cref{fig:just_right} is a VGG13 network with halved filter sizes on all layers and does not show this long-tail.
This property of saturation is also sensitive to the complexity of the problem.
We can see from \Cref{fig:complex_problem} that the same architecture is achieving generally lower saturation and a long-tail when encountered with a (simpler) problem with fewer classes.
When a network is overparameterized which can be seen in \Cref{fig:overparam}, the saturation stays high for longer, leading to considerably degraded performance.

The patterns described above are emerging during training, as can be seen in \Cref{fig:sat_3d}. However, especially underparameterized networks seem to start of with much higher saturation levels, making it possible to detect them early during training.

Analysing the saturation patterns of neural networks during training is indicative of how well the network can deal with the problem it is trained on. This could be useful for guiding architecture search and quickly ruling out under- and overparameterized architectures without relying on estimates of what baseline performance can be expected from a good model.

\section{Conclusions and Future Work}

\label{Conclusion and Future Work}
We presented an easy to interpret metric, which can be computed on-line and measured during training.
We showed that the metric is useful for understanding the information processing of any feed-forward neural network and determining whether a network is over- or underparamterized.
We further showed that saturation on a layer-wise level as well as averaged over the network structure is related to the model's performance as well as to the interaction between the difficulty of the dataset and the network's structure.
We demonstrated, that the sequence of layer-saturations inside the neural architecture follows a pattern, depending on how well the model is parameterized for the given problem.

For future work we would like to test the saturation metric against additional degrees of freedom in neural network training like batch-size, different regularization techniques and more complex problems like CIFAR100 or ImageNet.
Saturation also shows different properties of convergence than the training loss while overfitting, based on these findings we are also considering to develop an early stopping rule that is independent of the validation set.

\clearpage
\onecolumn

\appendix

\section{Additional plots from the experiments}

\subsection{Fully Connected Neural Network Experiments}

\begin{figure*}[htb!]
\centering
    \begin{subfigure}{0.5\textwidth}
          \centering
          \includegraphics[height=1.6in]{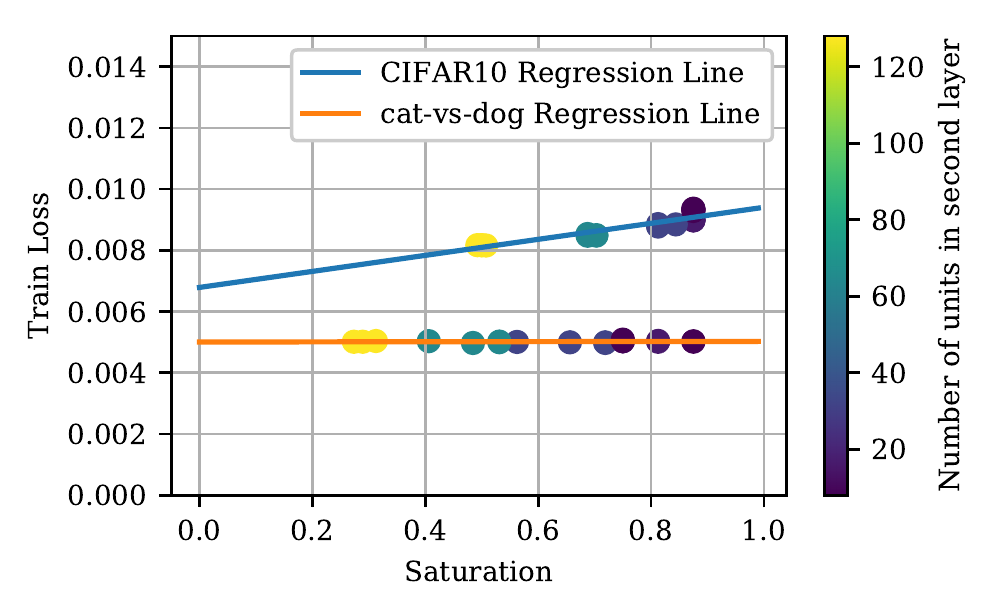}
          \caption[intrinsic_dims]{Test Loss}
          \label{fig:train_loss_dense}
    \end{subfigure}%
    ~
    \begin{subfigure}{0.5\textwidth}
          \centering
          \includegraphics[height=1.6in]{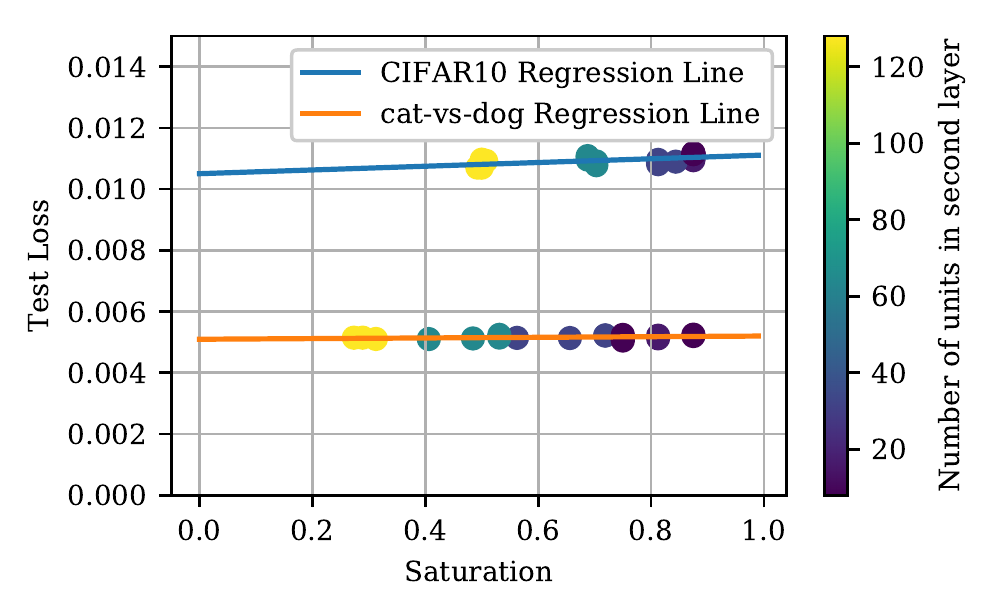}
          \caption[intrinsic_dims]{Train Loss}
          \label{fig:test_loss_dense}
    \end{subfigure}
    \caption[losses]{Train and test losses of all experiments conducted on CIFAR10 and \textit{cat-vs-dog}.}
    \label{fig:cnn_losses}
\end{figure*}

\subsection{CNN Experiments}

\begin{figure*}[htb!]
\centering
\includegraphics[scale=1.0]{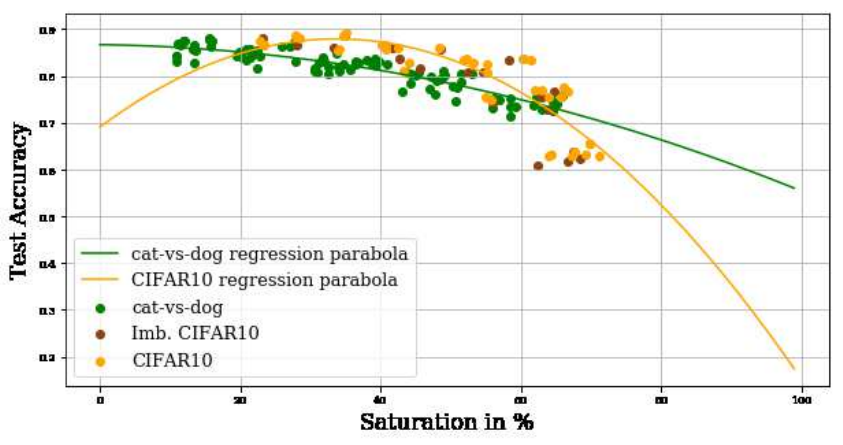}
\caption{The slope of the performance decrease becomes steeper for more complex problems. Note that random imbalances on CIFAR10 are not affecting the saturation abnormally.}
    \label{fig:scatter_conv_test_acc}
\end{figure*}

\begin{figure*}[htb!]
\centering
    \begin{subfigure}{.4\linewidth}
        \captionsetup{width=\columnwidth}
          \centering
          \includegraphics[width=\columnwidth]{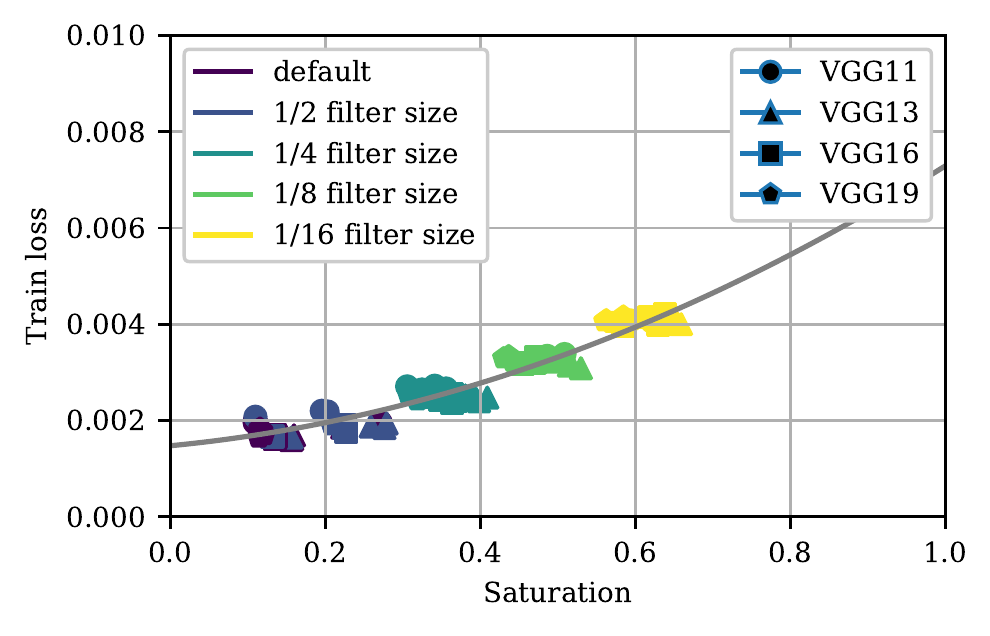}
          \caption[intrinsic_dims]{Cat vs Dog Train Loss}
          \label{fig:test_loss_cvd}
    \end{subfigure}
    \begin{subfigure}{.4\linewidth}
        \captionsetup{width=\columnwidth}
          \centering
          \includegraphics[width=\columnwidth]{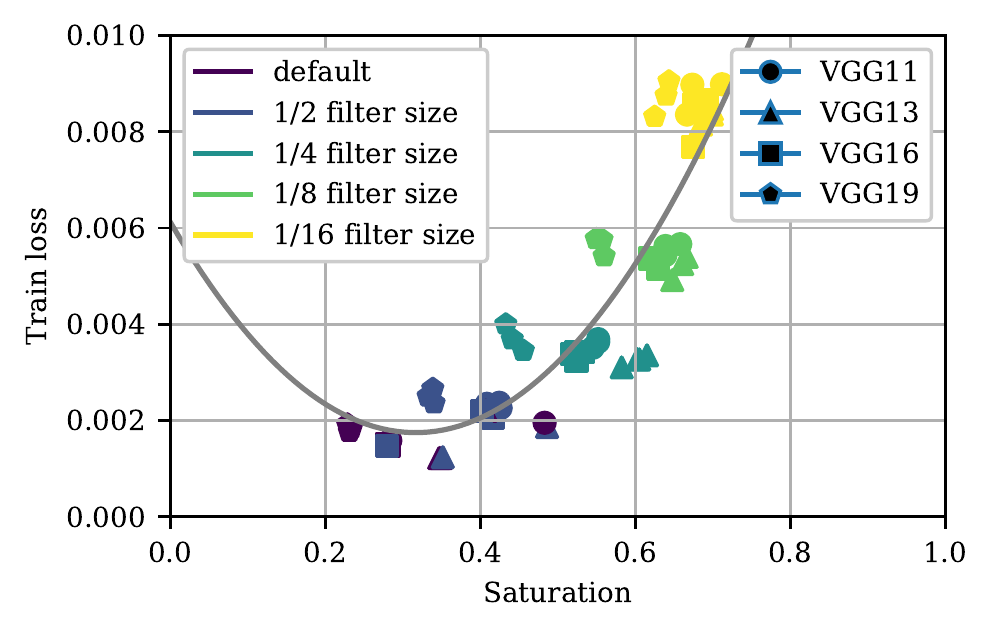}
          \caption[intrinsic_dims]{CIFAR10 Train Loss}
          \label{fig:test_loss_c10}
    \end{subfigure}
        \begin{subfigure}{.4\linewidth}
        \captionsetup{width=\columnwidth}
          \centering
          \includegraphics[width=\columnwidth]{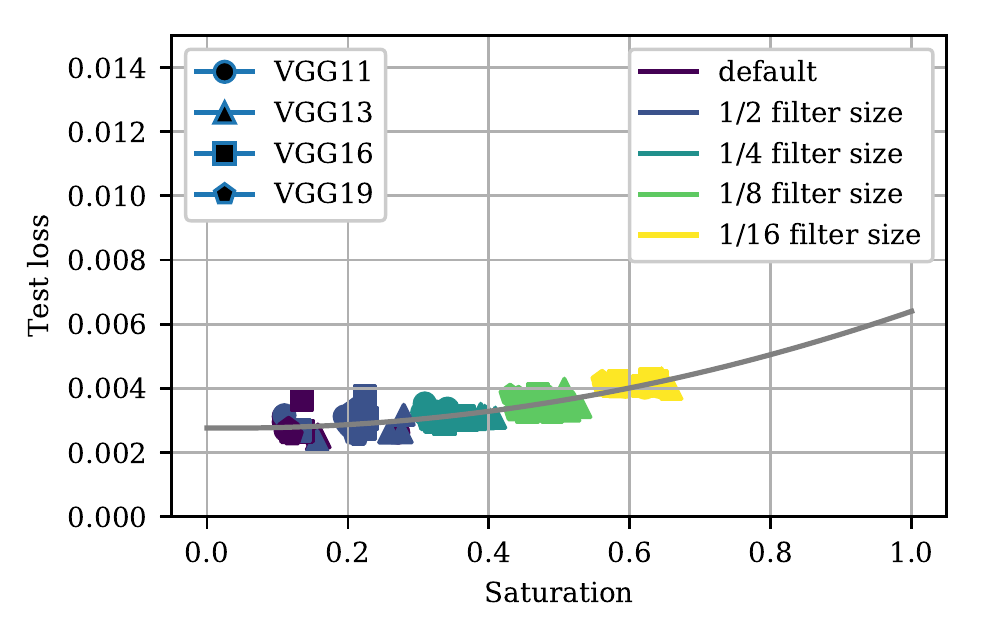}
          \caption[intrinsic_dims]{Cat vs Dog Test Loss}
          \label{fig:train_loss_cvd}
    \end{subfigure}
    \begin{subfigure}{.4\linewidth}
        \captionsetup{width=\columnwidth}
          \centering
          \includegraphics[width=\columnwidth]{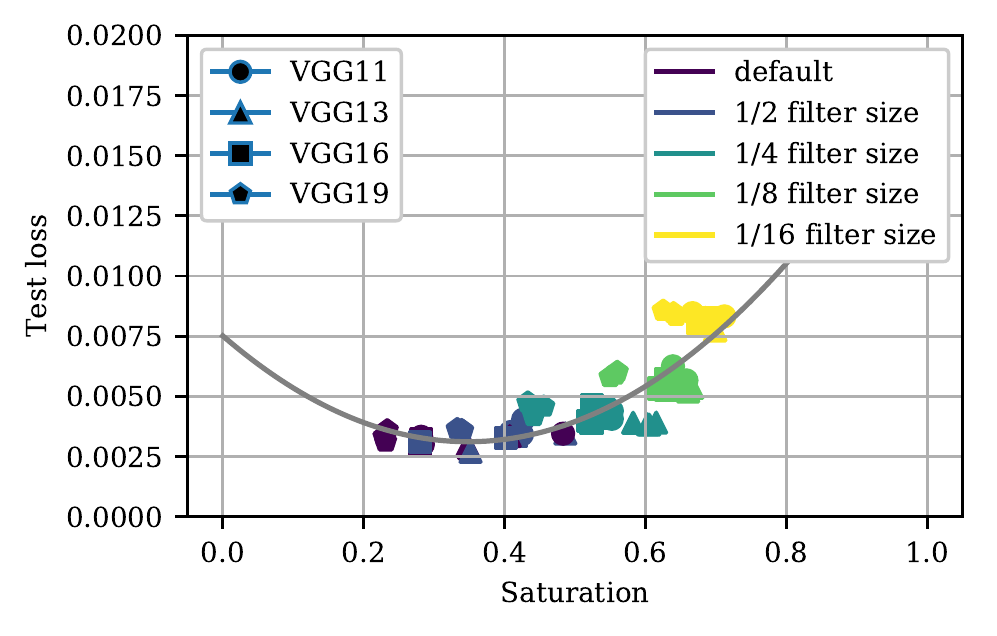}
          \caption[intrinsic_dims]{CIFAR10 Test Loss}
          \label{fig:train_loss_c10}
    \end{subfigure}
    \caption[losses]{Train and test losses of all experiments conducted on CIFAR10 and \textit{cat-vs-dog}. Note that the behavior is mostly similar to the test accuracy as seen in \Cref{fig:complex_problem}. }
    \label{fig:cnn_losses}
\end{figure*}

\begin{figure*}[t!]
\centering
\includegraphics[scale=0.6]{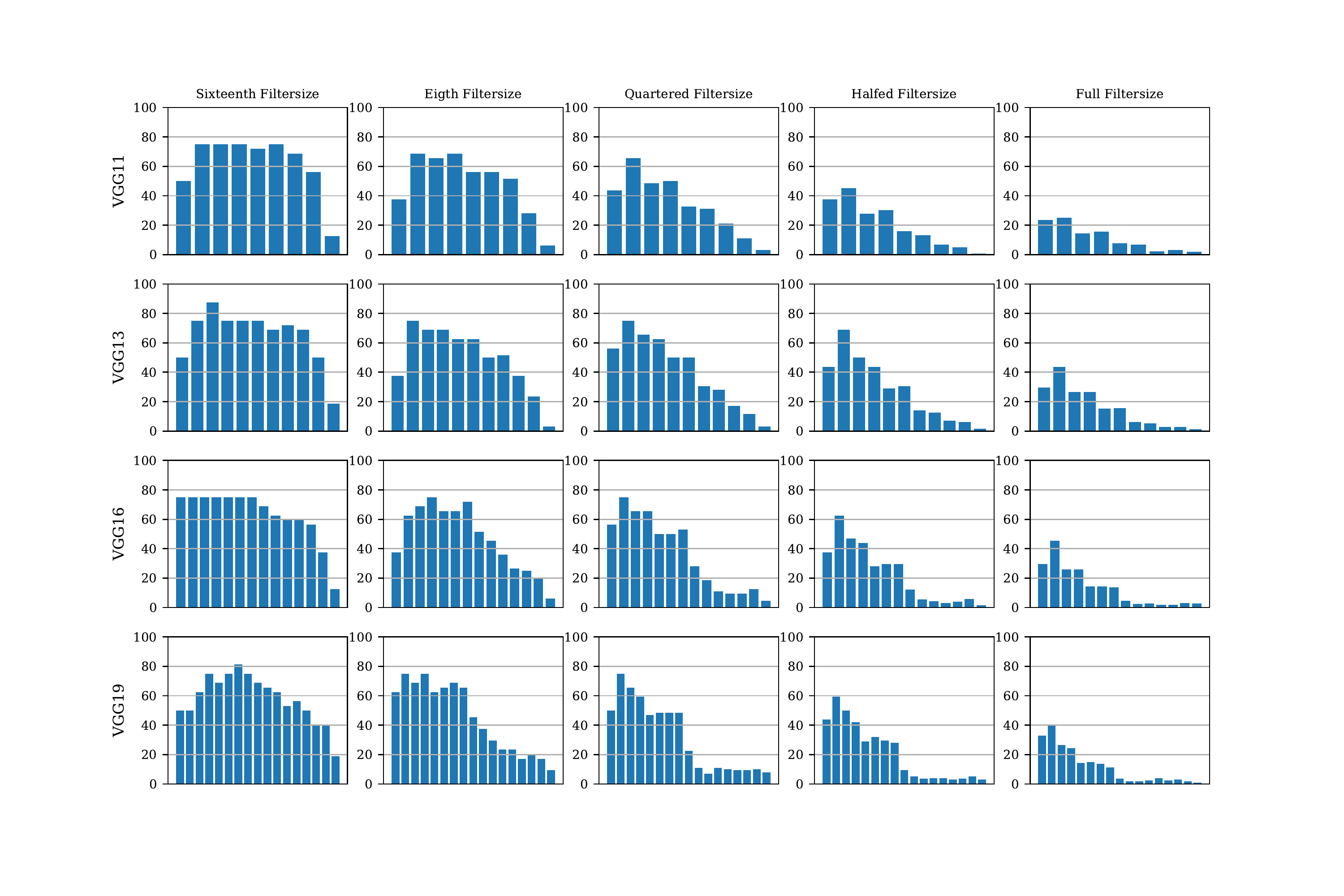}
\caption{Layer-wise saturations of all tested CNN architectures trained on \textit{cat-vs-dog} for 20 epochs. The layers are depicted on the x-axis in the same sequence as the information is propagated through the network at inference time. The y-axis describes the saturation value. Note the gradual change in the distribution while the filter size and depth increase.}
    \label{fig:collage_cd}
\end{figure*}

\begin{figure*}[t!]
\centering
\includegraphics[scale=0.6]{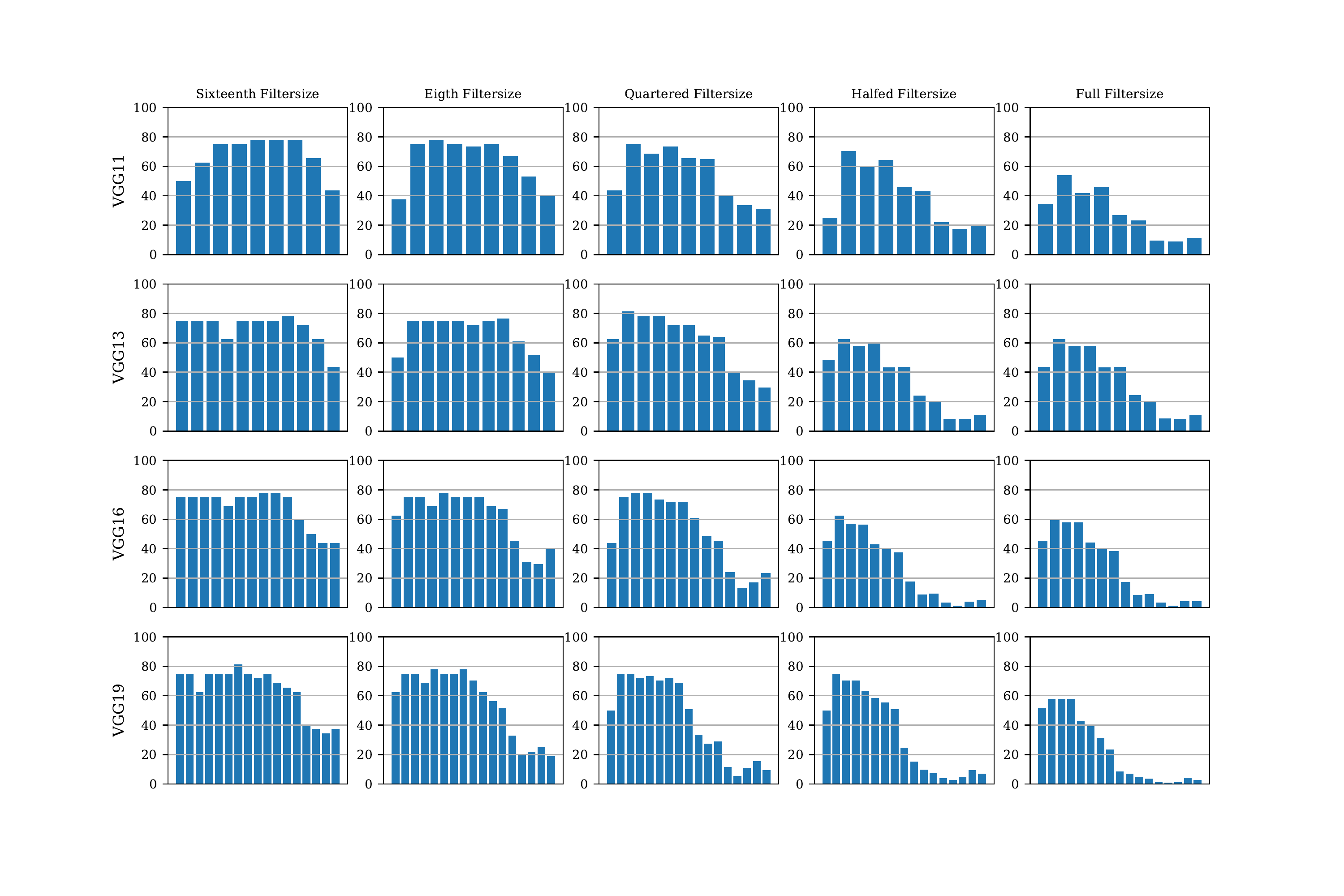}
\caption{Layer-wise saturations of all tested CNN architectures trained on CIFAR10 for 20 epochs. The layers are depicted on the x-axis in the same sequence as the information is propagated through the network at inference time. The y-axis describes the saturation value. Note the gradual change in the distribution while the filter size and depth increase.}
    \label{fig:collage_cf}
\end{figure*}


\begin{thebibliography}{10}

\bibitem{alexnet}
Alex Krizhevsky, Ilya Sutskever, and Geoffrey~E Hinton.
\newblock Imagenet classification with deep convolutional neural networks.
\newblock In F.~Pereira, C.~J.~C. Burges, L.~Bottou, and K.~Q. Weinberger,
  editors, {\em Advances in Neural Information Processing Systems 25}, pages
  1097--1105. Curran Associates, Inc., 2012.

\bibitem{zfnet}
Matthew~D. Zeiler and Rob Fergus.
\newblock Visualizing and understanding convolutional networks.
\newblock {\em CoRR}, abs/1311.2901, 2013.

\bibitem{keskar}
Nitish~Shirish Keskar, Dheevatsa Mudigere, Jorge Nocedal, Mikhail Smelyanskiy,
  and Ping Tak~Peter Tang.
\newblock On large-batch training for deep learning: Generalization gap and
  sharp minima.
\newblock {\em CoRR}, abs/1609.04836, 2016.

\bibitem{sensitivitygoogle}
Roman Novak, Yasaman Bahri, Daniel~A. Abolafia, Jeffrey Pennington, and Jascha
  Sohl-Dickstein.
\newblock Sensitivity and generalization in neural networks: an empirical
  study.
\newblock In {\em International Conference on Learning Representations}, 2018.

\bibitem{errorsurface}
Hao Li, Zheng Xu, Gavin Taylor, Christoph Studer, and Tom Goldstein.
\newblock Visualizing the loss landscape of neural nets.
\newblock In S.~Bengio, H.~Wallach, H.~Larochelle, K.~Grauman, N.~Cesa-Bianchi,
  and R.~Garnett, editors, {\em Advances in Neural Information Processing
  Systems 31}, pages 6389--6399. Curran Associates, Inc., 2018.

\bibitem{svcca}
Maithra Raghu, Justin Gilmer, Jason Yosinski, and Jascha Sohl-Dickstein.
\newblock Svcca: Singular vector canonical correlation analysis for deep
  learning dynamics and interpretability.
\newblock In I.~Guyon, U.~V. Luxburg, S.~Bengio, H.~Wallach, R.~Fergus,
  S.~Vishwanathan, and R.~Garnett, editors, {\em Advances in Neural Information
  Processing Systems 30}, pages 6076--6085. Curran Associates, Inc., 2017.

\bibitem{nin}
Min Lin, Qiang Chen, and Shuicheng Yan.
\newblock Network in network.
\newblock {\em CoRR}, abs/1312.4400, 2014.

\bibitem{vgg}
Karen Simonyan and Andrew Zisserman.
\newblock Very deep convolutional networks for large-scale image recognition.
\newblock {\em CoRR}, abs/1409.1556, 2014.

\bibitem{resnet}
Kaiming He, Xiangyu Zhang, Shaoqing Ren, and Jian Sun.
\newblock Deep residual learning for image recognition.
\newblock {\em CoRR}, abs/1512.03385, 2015.

\bibitem{CIFAR-10}
Alex Krizhevsky, Vinod Nair, and Geoffrey Hinton.
\newblock The cifar-10 dataset.
\newblock {\em online: http://www. cs. toronto. edu/kriz/cifar. html}, 55,
  2014.

\bibitem{inceptionv3}
Christian Szegedy, Vincent Vanhoucke, Sergey Ioffe, Jonathon Shlens, and
  Zbigniew Wojna.
\newblock Rethinking the inception architecture for computer vision.
\newblock {\em CoRR}, abs/1512.00567, 2015.

\end{thebibliography}
\end{document}